\documentclass[final,3p,12pt]{elsarticle}
\journal{Preprint}
\let\today\relax
\makeatletter
\def\ps@pprintTitle{%
    \let\@oddhead\@empty
    \let\@evenhead\@empty
    \def\@oddfoot{\footnotesize\itshape
         {Preprint} \hfill\today}%
    \let\@evenfoot\@oddfoot
    }
\makeatother
\usepackage{amsmath}
\usepackage{amsfonts}
\usepackage{amssymb}
\usepackage{amsthm}
\usepackage{bm}
\usepackage{mathtools}
\usepackage{indentfirst}
\usepackage{booktabs}
\usepackage[dvipsnames, table]{xcolor}
\usepackage{multirow}
\usepackage{subcaption}
\usepackage[hidelinks]{hyperref}
\usepackage{tikz}
\usepackage{array}
\usepackage[flushleft]{threeparttable}
\usepackage[export]{adjustbox}
\usepackage[column=0]{cellspace}
\tikzset{every picture/.style={line width=0.75pt}} 
\newtheorem{thm}{Theorem}[section]

\newtheorem{definition}[thm]{Definition}
\newtheorem{rem}[thm]{Remark}

\numberwithin{equation}{section}

\renewcommand{\appendix}{\par
  \setcounter{section}{0}
  \setcounter{subsection}{0}
  \gdef\thesection{\Alph{section}}
}

\newcommand{\norm}[1]{\left\Vert#1\right\Vert}
\newcommand{\abs}[1]{\left\vert#1\right\vert}

\newcommand{\set}[1]{\left\{#1\right\}}
\newcommand{\Real}{\mathbb R}
\usepackage{lineno}

\usepackage{ulem}

\begin{document}
\begin{frontmatter}
\title{Transformers as Neural Operators for Solutions of Differential Equations with Finite Regularity}

\author[1]{Benjamin Shih}
\author[1]{Ahmad Peyvan}
\author[2]{Zhongqiang Zhang}

\author[1]{George Em Karniadakis}
\affiliation[1]{organization={Division of Applied Mathematics, 182 George Street, Brown University},
    city={Providence},state={RI}, postcode={02912},country={USA}
            }
\affiliation[2]{organization={Department of Mathematical Sciences, Worcester Polytechnic Institute},
city={Worcester}, state={MA}, postcode={01609}, country={USA}}

\begin{abstract}
Neural operator learning models have emerged  
as very effective surrogates in data-driven methods
for partial differential equations (PDEs) across different applications from computational science and engineering. 
Such operator learning models not only predict
particular instances of a physical or biological system in real-time but also forecast classes of solutions corresponding to a distribution of initial and boundary conditions or forcing terms. 
DeepONet is the first neural operator model and has been tested extensively for a broad class of solutions, including Riemann problems. Transformers have not been used in that capacity, and specifically, they have not been tested for 
solutions of PDEs with low regularity.
 In this work, we first establish the theoretical groundwork that transformers possess the universal approximation property as operator learning models. 
 We then apply transformers to forecast solutions of diverse dynamical systems with solutions of finite regularity for a plurality of initial conditions and forcing terms. In particular, we consider three examples: 
 the Izhikevich neuron model, the tempered fractional-order
Leaky Integrate-and-Fire (LIF) model, and the one-dimensional Euler equation Riemann problem. For the latter problem, we also compare with variants of DeepONet, and we find that transformers outperform DeepONet in accuracy but they are computationally more expensive.
\end{abstract}



\end{frontmatter}

{\renewcommand\arraystretch{0.75}
 
\section{Introduction}
Operator learning using neural networks
 has been broadly explored and demonstrated to solve complex problems in computational science and engineering. 
 With a large amount of data, properly designed architectures of neural networks can be trained to solve large-scale computational problems efficiently.
 Once trained, the  neural networks
 can perform real-time forecasting as surrogate models, see e.g. \cite{lu2021learning,lu2022comprehensive}. 
 Also, they can be recycled for similar tasks with a little
 or without fine-tuning \cite{goswami2022Transferoperator}. 
 
 In this new paradigm of data-driven modeling, we formulate the underlying problems as nonlinear mappings between infinite- or high-dimensional input and output functions.
 The theoretical foundation of such operator learning using shallow networks has been laid in \cite{chen1995universal}, which has been extended to deep neural networks and implemented in \cite{lu2021learning}. 
%
Also, Fourier neural operators  (FNOs)\cite{li2020fourier,kovachki2021universal} and many consequent works are alternative architectures for operator learning. Thorough comparisons among DeepONets and FNOs have been reported in \cite{fno-don-compare}, where it has been shown that the performance among DeepONets and FNOs is comparable. 
The literature on operator learning has been quickly developing but here we use DeepONets as our baseline models.
 
Several improvements have been made for DeepONets.
We may use the physics-informed DeepONets, e.g., in  \cite{wang2021learning} and \cite{goswami2023physics}  to improve the accuracy. 
The accuracy of DeepONets may also be improved by adding a norm of the gradients, e.g. \cite{luo2023efficient}. 
Due to the structure of DeepONets, training {DeepONets} can be split into two steps as in DeepONets with POD \cite{lu2021comprehensive} or SVD \cite{Venturi2023svd-deeponet}: one can first perform singular value decomposition and use the eigenvectors scaled by the positive eigenvalues as the basis and emulate this basis to obtain trunk networks and subsequently train the branch networks with that basis. 
Further improvement in \cite{lee2023training}   can be made by first emulating the POD basis to obtain trunk networks and then approximating the POD coefficients to get the branch networks.
Discretization-independent DeepONets are proposed in \cite{zhang2023belnet} and 
the architecture of DeepONets is modified to accommodate mesh-based data in \cite{franco2023mesh-informed}. 
In \cite{DengSLZK22}, several architectures of DeepONets are presented by emulating efficient numerical methods for the underlying operators.

Transformers \cite{standardattention} have been successfully applied and modified for various differential equations and the literature along this avenue is quickly developing.
Here we name a few works, e.g.,
integral equations \cite{zappala2023neural},  dynamical systems \cite{GENEVA2022272},    fluid flows \cite{li2023-transformerPDE,li2024latent,liu2024mitigating,ovadia2023vito,ovadia2023ditto},  and inverse problems \cite{guo2022transformer}.    
However, neural operators and systems with \textit{rough solutions} such as hyperbolic problems with shocks have not been addressed.
In this work, we show that the transformer can outperform the state-of-the-art operator learning   \cite{riemannonet} (a variant of DeepONets) for Riemann problems. 

Theoretically, we prove that transformers as operator learning models are universal approximators in Section \ref{sec:uap-transformer}. 
In Section \ref{sec:problems}, we introduce three problems and explain how data is generated to train the operator learning models.
In these problems, we consider solution operators of two ordinary differential equations (ODEs) with jump forcing and high-pressure ratio for a classical one-dimensional shock tube problem. The solutions to the ODEs may have only bounded derivatives of order no more than one and the last problem may have at most bounded variations and jump discontinuities. Also, the second ODE is fractional, i.e., nonlocal in time and the problem has a long-term memory.
In Sections 4 and 5, we show that transformers outperform DeepONets obtained from the state-of-the-art training strategies for the problems in Section \ref{sec:problems}. 

The novelty and main contributions of the work are summarized as follows:
\begin{itemize}
    \item We establish the theoretical groundwork that transformers possess the universal approximation property as operator learning models. 
    \item We discover that transformers exhibit excellent performance for dynamical systems with long memory and rough solutions. 
    \item We compare operator learning models for the above problems. We also address the relative performance of two training optimizers: the classical Adam and the newly developed Lion. Moreover, we compare RiemannONet and transformer for Sod's shock tube problems. 
\end{itemize}
\section{Universal approximation of the transformer as an operator learning model}\label{sec:uap-transformer}

In the literature, it has been shown that the transformer is a universal approximator from sequence to sequence,  e.g.
\cite{cordonnier2020.RelationshipSelfAttentionConvolutional,pmlr-v202-takakura23a,Yun2020Are}.
Here we prove the universal approximation of a transformer (with a decoder) as an operator learning model.  

Let $\mathcal{G}$ be a continuous operator from 
$X$ to $Y$, where $X$ and $Y$ are both Banach spaces.
We show that the convergence is uniform for all 
$u\in U$, a compact subset of $X$. 
Since the transformer $T$ is in nature sequence to sequence, we need a proper decoder $\mathcal{D}_T$ and an encoder $\mathcal{E}_T$ such that 
$\mathcal{D}_T T (\mathcal{E}_T u)$ defines a continuous map from $X$ to $Y$.

\begin{thm}[Universal approximation of transformers for continuous operators] \label{thm:uap-transformer}
Suppose that $U$ is a compact subspace of 
$X$, defined on a compact set $K_{in}$ of $\mathbb{R}^{d_{in}}$. Then for any $\epsilon>0$, there exist some
integers $\mathsf{m},\mathsf{n}>0$ and encoder $\mathcal{E}_T: X\mapsto\Real^{\mathsf{m}}$, 
 and a transformer $T:\Real^{\mathsf{m}} \mapsto \Real^{\mathsf{n}} $ 
and decoder $\mathcal{D}_T: \Real^{\mathsf{n}} \mapsto Y$such that 
\begin{align*}
    \sup_{u \in U} \norm{G (u) -  \mathcal{D}_T T(\mathcal{E}_T u) }_Y<\epsilon.
\end{align*}
\end{thm}

The proof is similar to that in \cite{chen1995approximation,DengSLZK22,lanthaler2021error}. The basic  idea is that on a compact set of $X$, the uniform approximation of a continuous operator can be projected into uniform approximation in finite-dimensional Euclidean spaces.


For simplicity, we present the proof in the case of $X=C(K_{in})$ and $Y= C(K_{out})$ where $K_{out} \subset \mathbb{R}$ is compact. First, 
there exist encoders $\mathcal{E}_{\mathcal{G},i}$ and $\mathcal{E}_{\mathcal{G},i}$ such that 
\begin{align*}
    &\quad \sup_{u \in U} \norm{\mathcal{G}(u) -  \mathcal{E}_{\mathcal{G},o} \mathcal{G}(\mathcal{E}_{\mathcal{G},i} u) }_Y <  \frac{\varepsilon}{2}.
\end{align*}
This conclusion has been proved in many works, e.g. in  \cite{chen1995universal,DengSLZK22,lanthaler2021error}. 
Specifically, let 
$\set{x_j}_{j=1}^{\mathsf{m}}$ form a partition of $K_{in}$ and 
$\set{y_k}_{k=1}^{\mathsf{n}}$ form a partition of $K_{out}$. Then,  we may have the following approximation
\begin{equation}\label{eq:gu-approximation}
 \mathcal{G}(u) \approx \mathcal{G}(\mathcal{E}_{\mathcal{G},i}u) \approx  \mathcal{E}_{\mathcal{G},o}\mathcal{G}(\mathcal{E}_{\mathcal{G},i}u)=\sum_{k=1}^{\mathsf{n}} 
G( \sum_{j=1}^{\mathsf{m}}u(x_j)m_j(x))(y_k)e_k(y),   
\end{equation}
where $x_j \in K_{in}$ and
$y_k \in K_{out}$. Also,
$m_j(x)$/$e_k(y)$ is  a basis from a partition of  unity over $K_{in}$/$K_{out}$. 

Introduce the notation
$\mathbb{E}_{\mathsf{m}} u =
\mathbf{u} =\big(u(x_1),\ldots, u(x_{\mathsf{m}})\big)^\top$ where $u(x_j)$'s are  from 
\eqref{eq:gu-approximation}; $\mathbb{E}_{\mathsf{n}}v = \big(v(y_1),\ldots, v(y_{\mathsf{n}})\big)$ and $x_j$'s are  from 
\eqref{eq:gu-approximation}.
Then, by the compactness of $U\subset X$, there exists 
$M>0$ such that $u(x_j) \in [-M,M]$ for all $u\in U$ and $j$. 

Observe that the map from $\mathbf{u}$ to 
$\mathbb{E}_{\mathsf{n}}\mathcal{G}(\mathcal{E}_{\mathcal{G},i}u)$
is continuous.
By the universal approximation theorem of the transformer from sequence to sequence, e.g. in \cite{cordonnier2020.RelationshipSelfAttentionConvolutional,pmlr-v202-takakura23a,Yun2020Are}, there exists a transformer $T: \Real^{\mathsf{m}} \mapsto\Real^{\mathsf{n}}$ such that 
\[\sup_{\mathbf{u}\in [-M,M]^{\mathbf{m}}}\norm{T (\mathbf{u}) -\mathbb{E}_{\mathsf{n}}\mathcal{G}(\mathcal{E}_{\mathcal{G},i}u)}_{\ell^{\infty}}<\frac{\varepsilon}{2}.\]
Let $D_T : \Real^{\mathsf{n}} \mapsto Y$ be defined by 
$D_T (\cdot) = \sum_{k=1}^{\mathsf{n}}  (\cdot)_k e_k(y) $. Then 
\begin{align*}
   \norm{ \mathcal{E}_{\mathcal{G},o} \mathcal{G}(\mathcal{E}_{\mathcal{G},i} u) -  \mathcal{D}_T T(\mathbb{E}_{\mathsf{m}}\mathcal{E}_{\mathcal{G},i}u)}_Y
   &= \norm{\sum_{k=1}^n  (T(\mathbb{E}_{\mathsf{m}}\mathcal{E}_{\mathcal{G},i}u)-\mathbb{E}_{\mathsf{n}}\mathcal{G}(\mathcal{E}_{\mathcal{G},i}u))_k e_k(y)}_Y\\
   &\leq \norm{T (\mathbf{u}) -\mathbb{E}_{\mathsf{n}}\mathcal{G}(\mathcal{E}_{\mathcal{G},i}u)}_{\ell^{\infty}}<\frac{\varepsilon}{2}.
\end{align*}
Observe that  
$\sup_{u\in U}\norm{ \mathcal{E}_{\mathcal{G},o} \mathcal{G}(\mathcal{E}_{\mathcal{G},i} u) -  \mathcal{D}_T T(\mathbb{E}_{\mathsf{m}}\mathcal{E}_{\mathcal{G},i}u)}_Y \leq \sup_{\mathbf{u}\in 
[-M,M]^{\mathsf{m}}}\norm{T (\mathbf{u}) -\mathbb{E}_{\mathsf{n}}\mathcal{G}(\mathcal{E}_{\mathcal{G},i}u)}_{\ell^{\infty}}$.
Then, we obtain that, letting 
$\mathcal{E}_T=\mathbb{E}_{\mathsf{m}}\mathcal{E}_{G_i}$, 
\begin{align*}
\sup_{u \in X} \norm{\mathcal{G}(u) -  \mathcal{D}_T T(\mathcal{E}_{T} u) }&= \sup_{u \in X} \norm{\mathcal{G}(u) -  \mathcal{D}_T T(\mathbb{E}_{\mathsf{m}}\mathcal{E}_{G_i} u) }\\
    & \leq \sup_{u \in X} \norm{G (u) -  
    \mathcal{E}_{\mathcal{G},o} \mathcal{G}(\mathcal{E}_{\mathcal{G},i} u) }
        + \sup_{u \in X} \norm{\mathcal{E}_{\mathcal{G},o} \mathcal{G}(\mathcal{E}_{\mathcal{G},i} u) -  \mathcal{D}_T T(\mathbb{E}_{\mathsf{m}}\mathcal{E}_{\mathcal{G},i}u) }_Y\\
        &\leq \sup_{u \in X} \norm{G (u) -  
    \mathcal{E}_{\mathcal{G},o} \mathcal{G}(\mathcal{E}_{\mathcal{G},i} u) } + \sup_{\mathbf{u}\in 
[-M,M]^{\mathsf{m}}}\norm{T (\mathbf{u}) -\mathbb{E}_{\mathsf{n}}\mathcal{G}(\mathcal{E}_{\mathcal{G},i}u)}_{\ell^{\infty}}\\
    &< \frac{\varepsilon}{2}  + \frac{\varepsilon}{2}=\varepsilon.
\end{align*}
The universal approximation theorem is then proved.

Similar to the discussion in \cite{chen1995approximation,DengSLZK22}, we may present the following examples in Table \ref{tbl:uap-spaces}.
\begin{table}[!ht]
\begin{center}
\caption{Examples of functions spaces in Theorem \ref{thm:uap-transformer}. Here we also require the functions in the space of $U$  uniformly bounded in its norm such that $U$ is a compact subset of $X$.}
\label{tbl:uap-spaces}
\begin{tabular}{c !{\vrule width \lightrulewidth} c !{\vrule width \lightrulewidth} c  } 
    \toprule
  $X$  &  $Y$ & $U$    \\ 
    \midrule
   $X=C^0(K_{in})$  &  $Y=C^0(K_{out})$ & $C^{1}(K_{in})$    \\
   $X=L^2(K_{in})$  &  $Y=L^2(K_{out})$ & $H^1(K_{in})$ \\ 
   $X=L^1(K_{in})$  &  $Y=L^1(K_{out})$  & Bounded Variation \\
   \bottomrule
   \end{tabular}    
\end{center}    
\end{table}
In the following text, we are interested in $L^1$ spaces for $X$ and  $Y$.  The proof is the same as above except we replace the input 
$\mathbf{u} = (u(x_1),\cdots, u(x_{\mathsf{m}}))^\top$ with average values $\mathbf{u}$ = $(u_h (x_1 ),\ldots, 
u_h (x_\mathsf{m} ))$, where 
$	{u}_h (x_j)  = \int_{B(x_j,h)\cap K_{in}} u(t)\,dt/{\abs{B(x_j,h)\cap K_{in})}} $
	while $\abs{B(x_j,h)\cap K_{in}}$ is the volume of the ball $B(x_j,h)$ inside $K_{in}$, centered at $x_j$ with radius $h$ and the output with average values. See e.g. \cite{chen1993approximations} for more discussion. 
 Let's consider the solution operator of the following equation: $\partial_t u + \nabla \cdot f (u ) = 0$ for all $x\in \Real^d$ with the initial condition 
$u(x,0)=u_0(x)$. 
 Let $f \in C^1(\Real,\Real^d)$    satisfy the entropy condition.
 Then  we know from  \cite{holden2015front}
 that 
 the solution operator $\mathcal{G}$ 
 defined by $u=\mathcal{G}(u_0)$ is Lipschitz continuous: $\norm{\mathcal{G}(u_0)(t)-G(v_0)(t)}_{L^1}\leq \norm{u_0-v_0}_{L^1}$ for $ t>0$.
  	Here $u_0,v_0\in U = \{ u\in {\rm BV}(\Real^d)\cap L^1 (\Real^d): {\rm TV}(u) \leq M\}$ for some $M>0$,  where $\rm{BV}$	 is the space of functions of bounded variation and $\rm{TV}$ is the total variation. 
The conditions in Theorem \ref{thm:uap-transformer} are satisfied if $u_0$ is compactly supported and thus we can apply transformers to approximate this solution operator. 

\begin{rem}
Discussion about rates of approximation 
is beyond the scope of the paper
for difference spaces $X$ and $Y$ and subspaces. We refer interested readers for discussions in 
\cite{DengSLZK22,lanthaler2021error,Mhaskar23-localapproximation}.
\end{rem}
\section{Governing Equations and Data  Generation}\label{sec:problems}
We consider three problems: the Izhikevich neuron model, the tempered fractional-order Leaky Integrate-and-Fire (LIF) model, and the one-dimensional Euler equation Riemann problem.
\subsection{The Izhikevich Model}
The Izhikevich neuron model \cite{izhikevich2003simple} is described as
\begin{align}
    \frac{du}{dt} &= 0.04u^2 + 5u + 140 - v + I(t) \\
    \frac{dv}{dt} &= a(bu-v)
\end{align}
with the condition that when $u_{\operatorname{thres}}$ is some voltage threshold and $u \geq u_{\operatorname{thres}}$,
\[
\begin{cases}
    u \leftarrow c \\
    v \leftarrow v + d 
\end{cases}.
\]
In this system, $t$ is time, $u$ is the membrane recovery variable, $v$ is the membrane potential variable, $I(t)$ is the forcing spike function, and $a, b, c, d \in \Real$ are dimensionless.

We take $a=0.02,  b=0.25, c=-55,  d=0.05$ with a threshold value of $u_{\rm thres}=-64$. 
We consider the solution operator $\mathcal{G}: I(t) \to u$. 
Let the input current voltage spike $I$ vary between $0.5$ to $1.5$ such that the model is confined to the phasic bursting configuration, and the spiking location varies between $4$ to $30$ms in time.
The samples corresponding to spiking locations between $15$ and $19$ms are assigned to the test set, while the voltage spiking intensity is allowed to vary for the whole range between $0.5$ and $1.5$ for the test set. 
In particular, using a mesh of $401$ points across the time domain [0,100]ms, we generate $2,200$ training sample trajectories and $150$ testing sample trajectories. 



\subsection{Tempered Fractional LIF Model}
Traditionally, the LIF model is used as a neuron model that simulates different parts of the brain \cite{lif} and the integer-order model is formulated as:
\begin{equation}
    \tau\frac{dv}{dt} = -(v-v_{\operatorname{rest}}) +RI(t), 
    \label{eq: lif}
\end{equation}
where $\tau$ is known as a membrane time constant equivalent to $RC_m$ where $C_m$ is the membrane capacitance and $R$ is the membrane resistance. Further, $I(t)$ is the forcing spike function, and $v_{\operatorname{rest}}$ is the resting membrane potential.
The LIF model has been updated so that the derivative in \autoref{eq: lif} is of fractional order $\alpha$. We also investigate experiments with dependence on a temper order $\sigma$. That is, the tempered fractional model that we study is\begin{equation}
    \tau\prescript{\operatorname{C}}{a}{\mathcal{D}^{\alpha, \sigma}_t}[v](t) = -(v-v_{\operatorname{rest}}) +RI(t).
    \label{eq: frac-lif}
\end{equation}
Hence, the operator that we aim to approximate is:
\begin{equation}
    \mathcal{G}: (I(t), \alpha, \sigma) \mapsto v.
    \label{eq: lifop}
\end{equation}
In \autoref{eq: frac-lif}, $\prescript{\operatorname{C}}{a}{\mathcal{D}^{\alpha, \sigma}_t}$ refers to the tempered fractional Caputo derivative, defined as follows.
\begin{definition}[Tempered Caputo fractional derivative]
  The tempered fractional Caputo derivative with temper order $\sigma$ is defined as
  \begin{displaymath}
    \prescript{\operatorname{C}}{a}{\mathcal{D}^{\alpha, \sigma}_{t}}[u](t) = \frac{e^{-\sigma t}}{\Gamma (n-\alpha)} \int_a^t (t-s)^{n-\alpha-1} \frac{d^n}{ds^n}(e^{\sigma s} u(s)) ds.
  \end{displaymath}
  If $\sigma = 0$,  this reduces to the standard Caputo fractional derivative with fractional order $\alpha$.
\end{definition}
\subsubsection{Data Generation for the tempered fractional LIF Problem}
 We take $R = 5.1, C_m = 5\times 10^{-3}$, and $\tau=R\cdot C_m$, with $v_{\operatorname{rest}} = 0$. 
 For the tempered fractional LIF model, we use adaptive corrected $L1$-scheme solvers based on the work in \cite{fanhai}. 
 %
 We test three different cases of varying difficulty. In Case 1, the forcing term $I(t)$ is fixed with 3 spiking locations of the same amplitude. In Cases 2 and 3, the forcing varies with random vectors generated for location $t$ and spiking amplitude $I(t)$. Further details for each experiment case are given in the results and discussion. 
 
 In each case, we sample the fractional and temper parameter uniformly, e.g. for $\alpha$ we sample equispaced points on the interval $[0.11, 0.99]$ (we did not sample for $\alpha \leq 0.1$ due to numerical instability in the solver solutions) and similarly for $\sigma \in [0.11, 0.99]$.
 This is done for every case of the forcing and each batch of forcing would have varying solutions based on differing fractional and temper orders. 
 Each trajectory is obtained by solving the problem on an adaptive time mesh of $[0,1]$ second and thus the discretization for each experiment is different. 
 With these settings, we generate $200, 1500$, and $1500$ training sequence samples in the respective order of the experiment case, and $20, 200$, and $80$ testing sequence samples again in the order of the experiment case. 



\subsection{Riemann Problem}
Consider the Riemann problem for the one-dimensional Euler equations:

\begin{equation}
\frac{\partial \mathbf{U}(x,t)}{\partial t} + \frac{\partial \mathbf{F}(\mathbf{U})}{\partial x} = \mathbf{0}, \quad t \in [0, t_f], \quad \text{with} \quad \mathbf{U}(x,0) = \begin{cases}

\mathbf{U}_l & x \le x_s \\
\mathbf{U}_r & x > x_s
\end{cases}.
\label{eq:general}
\end{equation}
Here $t$ represents time, $x$ is the spatial coordinate, $\mathbf{U}$ is the vector of conservative variables, and $\mathbf{F}$ denotes the corresponding advective fluxes:
\begin{equation}
\mathbf{U} = \begin{pmatrix}
\rho\\
\rho u \\
\rho E
\end{pmatrix}^\top, \quad \mathbf{F}(\mathbf{U}) = \begin{pmatrix}
\rho u\\
\rho u^2 + p \\
u(\rho E + p)
\end{pmatrix} ^\top.
\label{eq:vec}
\end{equation}
In \autoref{eq:vec},  $\rho$ represents density, $p$ is pressure, $u$ is the velocity, and the total energy is expressed as:
\begin{equation}
\rho E = \frac{p}{\gamma - 1} + \frac{1}{2}u^2,\,\gamma =1.4.
\label{eq:enochem}
\end{equation}
Our goal is to determine the solution at a final time, $t=t_f$ with a given initial condition. Specifically, we consider the following operator 
\begin{align*}
    \mathcal{G}: p_l \to  \mathbf{U}(x,t_f),\end{align*}
where $p_l$ is the initial pressure for $x\leq x_s$. 
We investigate two cases with increasing learning difficulty. We parameterize the Riemann problem by fixing five states of the initial condition and varying the left initial pressure $p_l$ as discussed in \cite{riemannonet}. For the first case, we use an intermediate pressure ratio (IPR) for the initial conditions defined over $[-1,2]$. The initial conditions are selected as 
\begin{equation}
\left(\rho, u , p\right)=\begin{cases}
\left(2.0,0.0,p_l\right) & x_s \le 0.5 \\ 
\left(0.125,0.0,0.1\right) & x_s > 0.5, 
\end{cases}
\label{eq:low_p}
\end{equation}
where $p_l\in [50.0,100.0]$. The profiles of primitive variables e.g. $\rho$, $u$, and $p$, are obtained at $t_f=0.1$. For the second case, we employ an extremely high-pressure ratio (HPR). We solve Equation \eqref{eq:general} within the spatial domain $x\in[-20, 20]$. For this scenario, the initial conditions are chosen as
\begin{equation}
\left(\rho, u , p\right)=\begin{cases}
\left(2.0,0.0,p_l\right) & x_s \le -10 \\ 
\left(0.001,0.0,1.0\right) & x_s > -10, 
\end{cases}
\label{eq:high_p}
\end{equation}
where $p_l \in [10^9,10^{10}]$. Using the analytical method, we obtain the exact results at $t_f=0.0001$. We choose 500 equally-spaced values of $p_l$ and randomly allocate 400 cases for training and 100 cases for testing.
\section{Methodology}
In this section, we provide the details on the transformer architecture used for the operator learning problems and the loss and training optimizers.

\subsection{The Attention Mechanism}
%
In our experiments, we use the Fourier type-linear attention in \cite{Cao2021transformer}:
\begin{equation}
    \operatorname{Fourier Attention}(Q, K, V) = \frac{1}{n}(\widehat{Q}\widehat{K}^\top)V
\end{equation}
where $\widehat{\cdot}$ denotes layer normalization and the three matrices $Q, K, V$ are batches of sets of vectors: the query vectors $\{q_i\}$, key vectors $\{k_i\}$, and value vectors $\{v_i\}$. 
 We also use the cross-attention mechanism to accommodate the query points different from the input sensor points:
\begin{equation}
z_s(y_k) = \sum_{i=1}^d \frac{\sum_{j=1}^n k_i(x_j) v_s(x_j) }{n} q_i(y_k) 
\end{equation}
where $\{y_k\}$ are discretized points on the output domain and $\{x_j\}$ are the input grid points. Here, similar to the previous interpretation that the columns of $Q, K, V$ matrices contain vector representation of learned bases, $\{k_i(\cdot), v_i(\cdot), q_i(\cdot)\}$ represent three sets of basis functions as discussed in \cite{OFormer}. We can draw a parallel between cross attention and DeepONet, where $q_i(y_k)$ is the output of the trunk net and the inner summation $\sum_{j=1}^n k_i(x_j)v_s(x_j)$ represents the output of the branch net.

\subsection{Architecture}
Following the traditional sequence transduction models in \cite{standardattention}, we use the transformer in the form of an encoder-decoder structure depicted in \autoref{fig: architecture}, where the encoder maps the inputs into a latent representation from which an output sequence $\mathcal{G}_\theta[f_k; \Pi]$ is generated by the decoder.
In particular, mirroring the structure of the original DeepONet, the encoder structure consists of two different components: one which takes in the spatial or temporal domain discretization and embeds into the latent dimension, while the other parameters, forcing, terms, etc. are embedded using attention by another encoder component.

In addition, this attention encoder takes in any problem parameter set $\Pi$; in the LIF problem, these are the fractional and temper parameters $\alpha$ and $\sigma$. As previously described, the fully connected layer component of the encoder is used to encode the spatial or temporal query points, so either a sequence $\{t_n\}$ or $\{x_n\}$.
These encodings are then combined in the latent dimension by the cross attention mechanism originally discussed in \cite{chen2021crossvit}; the work of \cite{OFormer} inspires this architecture in which PDE benchmarks are explored using the transformer. 
For all experiments, we employ the same transformer structure, varying only the input dimension. 
In particular, for the baseline results, a model with an encoder embedding dimension for the sequence is 96 and 4 layers of attention are used. For the decoder, a decoding depth of 3 is used. Before the Fourier attention, embedding is done with fully connected layers; fully connected layers with GeGELU activation functions \cite{gegelu} are also applied after the cross-attention in the decoder to return the embedding into physical space.
\begin{figure}[tb!]
\centering
\begin{tikzpicture}[x=0.75pt,y=0.75pt,yscale=-1,xscale=1]
\draw[fill=blue!30!white]  (210,200) -- (55,270) -- (55,55) -- (210,120) -- cycle;
\draw[fill=red!30!white] (255,120) -- (300,80) -- (300,245) -- (255,200) -- cycle;

\draw [fill=yellow!30!white](70, 165) rectangle ++(25,80);
\draw (40,214.6) -- (70,214.6); 
\draw [shift={(70,214.6)}, rotate = 180] [color={rgb, 255:red, 0; green, 0; blue, 0 } ][line width=0.75]    (10.93,-3.29) .. controls (6.95,-1.4) and (3.31,-0.3) .. (0,0) .. controls (3.31,0.3) and (6.95,1.4) .. (10.93,3.29); 

\draw [fill=yellow!30!white](70, 80) rectangle ++(25,80);
\draw (40,115.6) -- (70,115.6); 
\draw [shift={(70,115.6)}, rotate = 180] [color={rgb, 255:red, 0; green, 0; blue, 0 }  ][line width=0.75]    (10.93,-3.29) .. controls (6.95,-1.4) and (3.31,-0.3) .. (0,0) .. controls (3.31,0.3) and (6.95,1.4) .. (10.93,3.29); 

\draw [fill=yellow!30!white](120, 105) rectangle ++(25,80);

\draw (95,125) -- (120,125); 

\draw (145,140) -- (158,140); 

\draw (95,205) -- (158,205); 

\draw (158,140) -- (158, 205);

\draw (158,157.6) -- (170,157.6); 

\draw [fill=yellow!30!white](170, 120) rectangle ++(25,80);

\draw (195,157.6) -- (229,157.6); 

\draw [fill=Orange!30!white](220, 120) rectangle ++(25,80);

\draw (245,157.6) -- (265,157.6); 

\draw [fill=yellow!30!white](265, 120) rectangle ++(25,80);
\draw (290,157.6) -- (330,157.6); 
\draw [shift={(330,157.6)}, rotate = 180] [color={rgb, 255:red, 0; green, 0; blue, 0 }  ][line width=0.75]    (10.93,-3.29) .. controls (6.95,-1.4) and (3.31,-0.3) .. (0,0) .. controls (3.31,0.3) and (6.95,1.4) .. (10.93,3.29); 

\draw (88,80) node [anchor=north west][inner sep=0.75pt]  [font=\tiny, rotate=-90, align=center] {Fourier Attention};
\draw (88,170) node [anchor=north west][inner sep=0.75pt]  [font=\tiny, align=center, rotate=-90] {Fully Connected};
\draw (138,107) node [anchor=north west][inner sep=0.75pt]  [font=\tiny, align=center, rotate=-90] {Fully Connected};
\draw (188,123) node [anchor=north west][inner sep=0.75pt]  [font=\tiny, rotate=-90, align=center] [align=left]{Cross Attention};
\draw (237,118) node [anchor=north west][inner sep=0.75pt]  [font=\tiny,rotate=-90] [align=left] {Latent Embedding};
\draw (0,107.5) node [anchor=north west][inner sep=0.75pt]{$f_k, \Pi$};
\draw (25,208.5) node [anchor=north west][inner sep=0.75pt]{$t$};
\draw (283,123) node [anchor=north west][inner sep=0.75pt]  [font=\tiny, rotate=-90, align=center] [align=left]{Fully Connected};
\draw (335,148) node [anchor=north west][inner sep=0.75pt]   {$\mathcal{G}_\theta[f_k; \Pi]$};

\draw (250,260) node [anchor=north west][inner sep=0.75pt] [align=left] {Decoder};
\draw (110,260) node [anchor=north west][inner sep=0.75pt] [align=left] {Encoder};
\end{tikzpicture}
\caption{Transformer architecture to approximate an operator $\mathcal{G}$. Inputs to the encoder are the forcing or initial condition ($f_k$ or $p_l$) in addition to the operator parameters $\Pi$ and domain discretizations $\{x_n\}$ or $\{t_n\}$. The output is the model's operator approximation with the input $f_k$ or $p_l$ and parameter set $\Pi$, $\mathcal{G}_\theta[f_k; \Pi]$.}
\label{fig: architecture}
\end{figure}
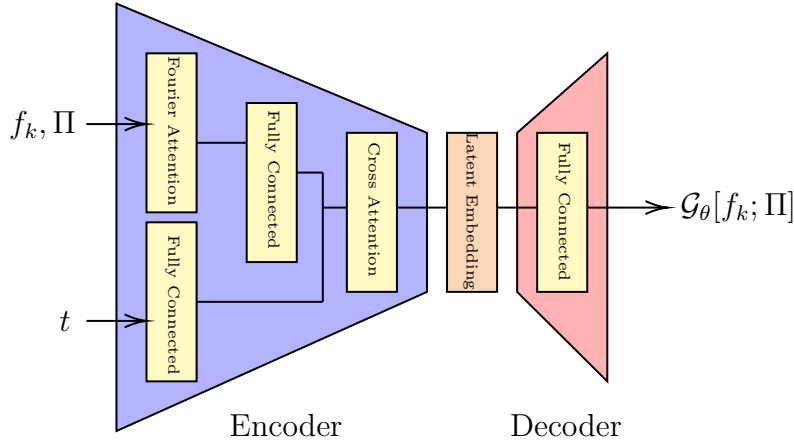
\subsubsection{Loss and Optimization}
To train the model, we minimize the following loss according to the relative $\ell_2$ norm:
\begin{equation}
    \mathcal{L}(\theta) = \frac{1}{m}\sum_{k=1}^m \frac{\|\mathcal{G}_\theta[f_k; \Pi] - \mathcal{G}[f_k; \Pi] \|_{\ell^2}}{\| \mathcal{G}[f_k; \Pi]\|_{\ell^2}}
\end{equation}
where $\mathcal{G}_\theta$ is the model prediction of some input $f_k$, and $\mathcal{G}$ is the corresponding ground truth solution for the same input. 
Here $\|f\|_{\ell^2} = \left(\frac{1}{n}\sum_{i=1}^n |f(t_i)|^2 \right)^\frac{1}{2}$. 
We used the Adam optimizer \cite{adam} for its prevalence in problems with many parameters and computational efficiency in combination with the learning rate scheduler known as the ``1cycle'' policy, as described in \cite{onecyclelr}. Additionally, we conducted experiments using the recently proposed Lion optimizer \cite{lion-optimizer}. 
\section{Results}
In this section, we present specific parameter settings and comparisons among the experimental results from different models.

We follow the guidance in \cite{lion-optimizer} for Lion: $5$ to $10$ times smaller learning rates than those for  Adam and larger batch sizes. 
We run the Lion optimizer with less than half epochs than in Adam. 

\subsection{Example 1: the Izhikevich Model}
We describe in \autoref{table:izhik-res} the training configurations and results using the Adam and Lion optimizer. We present some testing results in  \autoref{fig: izhik} where the Adam optimizer is used. In \autoref{table:izhik-res}, $n$ describes the number of points in the domain discretization. 
In \autoref{fig: izhik}, we observe that 
the solutions have sharp transitions and are oscillatory and both Adam and Lion optimizers 
lead to losses of the same magnitude. 
Here the Lion slightly outperforms the Adam while the standard deviation from the Lion model is an order smaller.
We note that the vanilla DeepONet did not work for this example.

\begin{table}[tb!]
    \caption{Experiment details and relative $\ell_2$ error for the Izhikevich model experiments. The loss $\mathcal{L}(\theta)$ reported is the mean and standard deviation of the relative $\ell_2$ losses across 10 runs.}
    \centering
    \scalebox{0.73}{
    \begin{tabular}{c|cccccc!{\vrule width \lightrulewidth}c} 
        \toprule
        \multicolumn{7}{c!{\vrule width \lightrulewidth}}{Experiment settings} & Results \\ 
        \midrule
            Model & $n$ & Spike Intensity & Spike Location & Learning Rate & Batch Size & Iterations & $\mathcal{L}(\theta)$ \\ 
        \midrule
        Adam & \multirow{2}{*}{$401$} & \multirow{2}{*}{$[0.5,1.5]$} & \multirow{2}{*}{$[4, 30]$} & $1.0 \times 10^{-3}$ & $64$ & $2.0 \times 10^5$ & $0.61 \pm 0.275$ \\
        Lion & & & & $1.0\times10^{-4}$& $500$ & $2.0 \times 10^{4}$& $0.49 \pm 0.080$ \\
        \bottomrule
    \end{tabular}}
    \label{table:izhik-res}
\end{table}

\begin{figure}[tb!]
    \centering
    \includegraphics[width=0.95\textwidth]{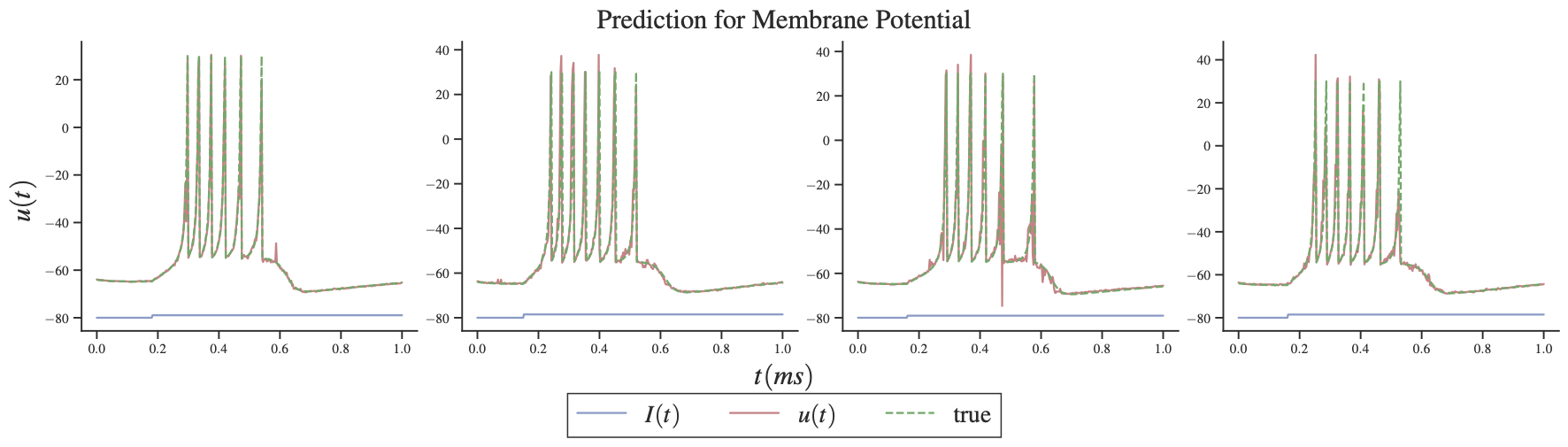}
    \caption{Adam model predictions for the Izhikevich experiment for varying spiking term intensities and discontinuity locations. The dashed green line represents the solver solution while the red line $u(t)$ is the model prediction, and the blue line is the spiking term $I(t)$.}
    \label{fig: izhik}
\end{figure}

\subsection{Example 2: Tempered Fractional LIF Model}
In this example, we consider the following solution operators defined by 
\begin{itemize}
    \item Case 1. 
    $V = \mathcal{G} (\alpha)$ and set $\sigma=0$;
\item Case 2. 
    $V = \mathcal{G} (I(t),\alpha)$ and set $\sigma=0$;
\item Case 3. 
    $V = \mathcal{G} ( I(t), \alpha, \sigma)$, same  as in \autoref{eq: lifop}. 
\end{itemize}
%
The fractional LIF model is nonlocal in time,  
and the memory of the system may depend on 
the tempered index $\sigma$. 
This model can be challenging to operator learning models for being non-Markovian and the spiking forcing. 
For example, we observe that even with a learning rate $1.0 \times 10^{-4}$, the Adam optimizer fails and hence we use a smaller learning rate of $5.0 \times 10^{-5}$ for Case 3, which presents the most difficulty as shown in \autoref{table:lif-res}.

In \autoref{table:lif-res}, we observe that for Case 1, the transformer with both optimizers leads to small losses with small standard deviations.
However, the transformer struggles in Cases 2 and 3, where the forcing terms have varying spikes at different locations. 
%
From \autoref{fig: lif-c2}, we observe that the transformer can capture the locations of the spikes. However, it cannot well predict the shape and amplitude of the solution membrane potential. 
Each prediction for membrane potential spikes in the right location, but the shape and amplitude of the potential curve do not exactly match the true curve. 

\begin{table}[tb!]
    \caption{Experiment details and relative $\ell_2$ error for each case of the tempered fractional LIF model for the two different models. The intervals given in the $\alpha$ and $\sigma$ columns represent the interval over which the parameter was sampled, while the values given in the $n$ columns are the numbers of points over which forcing terms and solutions were discretized. The loss $\mathcal{L}(\theta)$ reported is the mean and standard deviation of the relative $\ell_2$ loss across 10 runs.}
    \centering
    \scalebox{0.73}{
    \begin{threeparttable}
    \begin{tabular}{c|cc!{\vrule width \lightrulewidth}cccccc!{\vrule width \lightrulewidth}c} 
        \toprule
        \multicolumn{9}{c!{\vrule width \lightrulewidth}}{Experiment settings} & Results \\ 
        \midrule
            Model & Experiment & Case & $\alpha$  & $\sigma$ & $n$ & Learning Rate & Batch Size & Iterations & $\mathcal{L}(\theta)$ \\ 
        \midrule
        \parbox[t]{2mm}{\multirow{3}{*}{\rotatebox[origin=c]{90}{Adam}}} & \multirow{3}{*}{LIF} & $1$ & \multirow{3}{*}{$[0.11, 0.99]$} & --- & $204$ & \multirow{2}{*}{$1.0\times10^{-4}$}& \multirow{3}{*}{$32$} &\multirow{3}{*}{$5.0 \times 10^{5}$} & $7.79 \pm 1.14  (10^{-5})$ \\
            & & $2$ & & --- & $200$ & & & & $0.03 \pm 0.01\tnote{*}$\\
            & & $3$ & & $[0.11, 2]$ & $314$ & $5.0 \times 10^{-5}$ & & & $0.25 \pm 0.26\tnote{*}$\\
        \midrule
        \parbox[t]{2mm}{\multirow{3}{*}{\rotatebox[origin=c]{90}{Lion}}} & \multirow{3}{*}{LIF} & 1 & \multirow{3}{*}{$[0.11, 0.99]$} & --- & $204$ & \multirow{3}{*}{$1.0\times10^{-5}$} & $200$ & \multirow{3}{*}{$2.0 \times 10^{5}$} & $2.09 \pm 0.324 (10^{-4})$ \\
            & & $2$ & & --- & $200$ & & \multirow{2}{*}{$1000$} & & $0.10 \pm 0.02$ \\
            & & $3$ & & $[0.11, 2]$ & $314$ & & & & $0.06 \pm 0.02$\\
        \bottomrule
    \end{tabular}
    \begin{tablenotes}
        \item[*] Denotes losses whose reported mean and standard deviation were taken across ensemble models that converged---due to the Adam optimizer instability on these problems, some model losses exploded and were excluded.
    \end{tablenotes}
    \end{threeparttable}
    }
    \label{table:lif-res}
\end{table}

\begin{figure}[tb!]
    \centering
        \includegraphics[width=0.95\textwidth]{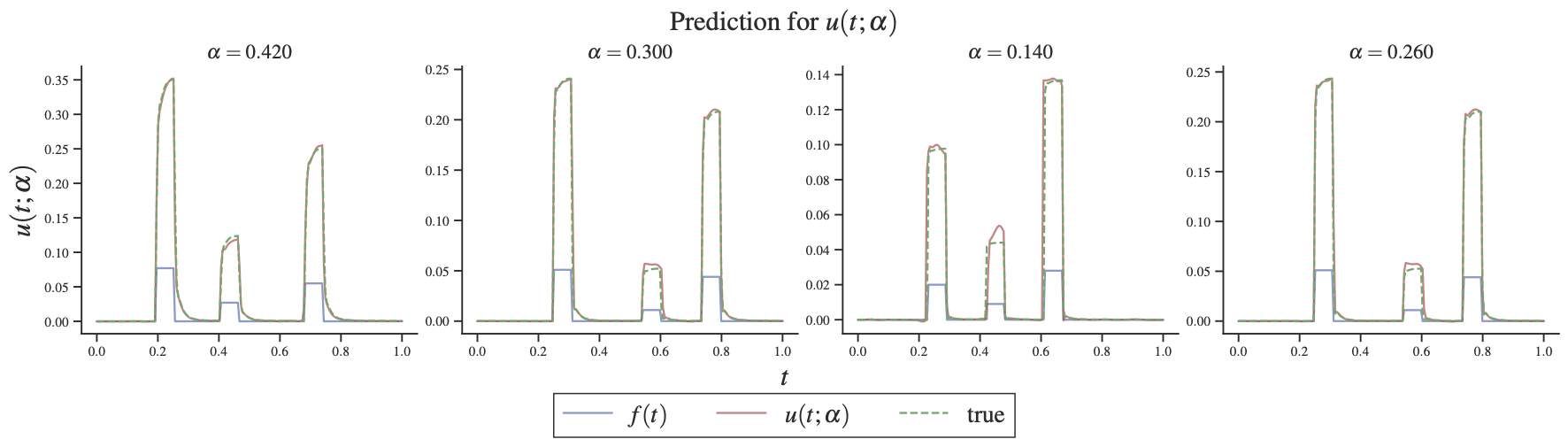}
    \caption{Lion model predictions for Case 2 of the LIF experiments varying the fractional order $\alpha$ and the spiking intensity and discontinuity location. The dashed green line is the solver solution while the red line $u(t; \alpha, \sigma)$ is the model prediction. The blue line is the spiking term.}
    \label{fig: lif-c2}
\end{figure}

In Case 3, we test the capability of the transformer when we vary the fractional and tempering orders and spiking force. 
The Adam optimizer exhibits final $l_2$ errors on the order of $10^{-1}$ while the Lion optimizer exhibits an error of one order of magnitude smaller. From \autoref{fig: lif-c3} where we use the Lion optimizer, we observe that the transformer predicts extremely well in most cases, except the left plot with slightly unresolved frequency, i.e., misalignment between the prediction (red solid line) and truth (green dashed line), when $\alpha = 0.5$ and $\sigma = 0.3$.

\begin{figure}[tb!]
    \centering
    \includegraphics[width=0.95\textwidth]{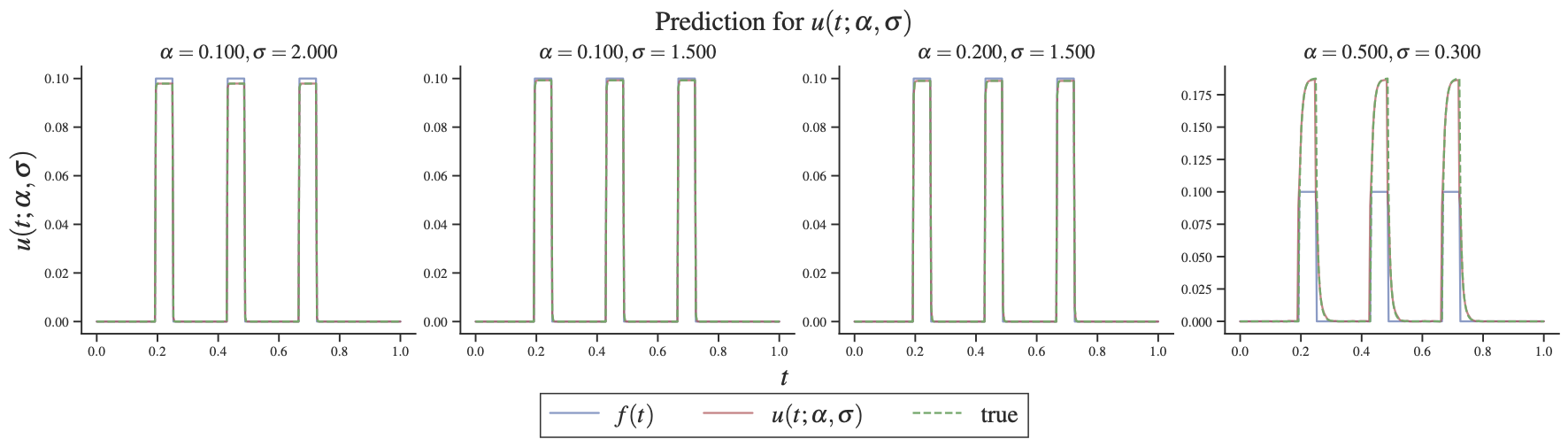}
    \caption{Lion model predictions for Case 3 of the LIF experiments varying both the temper order $\sigma$ and the fractional order $\alpha$. The dashed green line is the solver solution while the red line $u(t; \alpha, \sigma)$ is the model prediction. The blue line is the spiking term.}
    \label{fig: lif-c3}
\end{figure}

In this example, the transformer demonstrates the capability to learn the temporal dynamics and memory effects. 
We train the transformer with the Lion optimizer with polynomial learning rate decay for approximately half the epochs as the Adam model with 1cycle learning scheduler.
We observe the 1cycle learning rate scheduler allows the Adam model to achieve a decent loss on the order of $10^{-2}$ after about 100,000 epochs. However, the loss spikes up multiple times and eventually the training fails.
We observe that the loss from the Lion optimizer decays sharply and roughly monotonically with noticeably much less oscillation in the training loss. 
The accuracy from the Lion is an order higher than that achieved by the Adam model. 
Additionally, we observe from our experiments that the transformers trained with the  Lion optimizer exhibit better stability
while those trained with the Adam optimizer sometimes fail.

\begin{figure}[tb!]
    \centering
    \begin{subfigure}{0.49\textwidth}
      \centering
      \includegraphics[width=\linewidth]{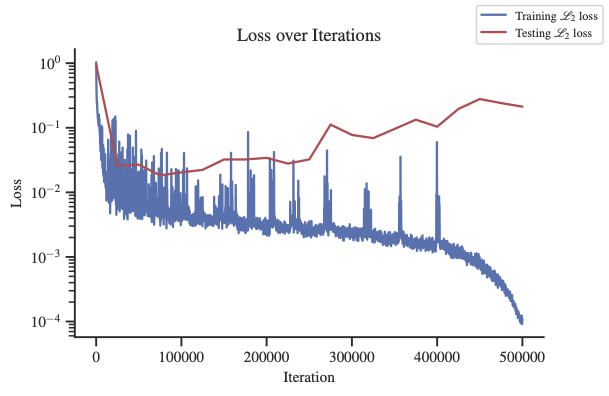}
      \caption{Adam/1cycle loss curve}
      \label{fig:sm-adam-loss}
    \end{subfigure}
    \begin{subfigure}{0.49\textwidth}
      \centering
      \includegraphics[width=\linewidth]{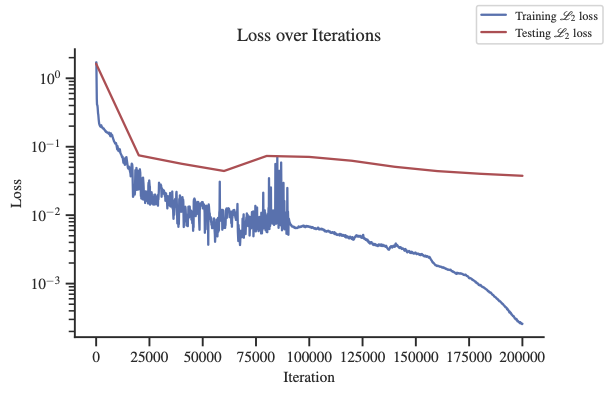}
      \caption{Lion/polynomial loss curve}
      \label{fig:sm-lion-loss}
    \end{subfigure}
    \caption{Comparison between the loss curves of the two models from Case 3 of the LIF experiments varying the fractional order $\alpha$ and the temper order $\sigma$. The loss curve for the Adam model is shown on the left (\autoref{fig:sm-adam-loss}) while the loss curve for teh Lion model is shown on the right (\autoref{fig:sm-lion-loss}).}
    \label{fig:sm-losses}
\end{figure}

\subsection{Example 3: Riemann Problem}
In this example, we compare the transformer and the RiemannONet \cite{riemannonet}. 
We test two different cases of the Riemann problem: intermediate-pressure ratio (IPR) and high-pressure ratio (HPR) Sod problems. To make fair comparisons, we follow the same experimental setup of RiemannONet and use the Lion optimizer only.
In the comparisons, we refer to the DeepONet in 
\cite{riemannonet} as 2-step DeepONet or 2-step Rowdy, where DeepONets are 
trained with the two-step strategy in \cite{lee2023training} and Rowdy activation functions.  

We report the mean and standard deviation of $\ell_2$ error on the testing set of an ensemble of 10 model runs. The runs are with different weight initialization and the mean training time for the ensemble is reported in \autoref{table: riemann-res}. 
The transformer outperforms the best RiemannONet in almost all experiments, except in the case of HPR. 
In some experiments (e.g. IPR $p$), the transformer can achieve less than half the relative error incurred by RiemannONet.

\begin{table}[tb!]
    \caption{Riemann problem: Relative $\ell_2$ norms with mean and standard deviation obtained using 10 runs. The $\ell_2$ norm of the error is calculated over the entire testing dataset for each of the profiles: density, velocity, and pressure.}
    \centering
    \scalebox{0.85}{
    \begin{tabular}{c|cccc} 
        \toprule
    Cases &$L_2(\rho)$ \%&$L_2(u)$ \%&$L_2(p)$ & Time (min) \\
    \midrule
    IPR (2 step Rowdy) & $0.33\pm0.027$ & $0.86\pm0.071$ & $0.20\pm0.030$& $33.06$\\
    IPR(Transformer) & $\mathbf{0.24 \pm 0.053}$& $\mathbf{0.58 \pm 0.141}$& $\mathbf{0.09 \pm 0.026}$& $343.343$\\
    \midrule
    HPR (2 step Rowdy) & $\mathbf{0.66\pm0.093}$   & $3.39\pm0.104$ & $2.86\pm1.680$&$26.98$\\
    HPR (Transformer) & $1.42 \pm 0.132$ & $\mathbf{2.27 \pm 0.338}$ & $\mathbf{2.23 \pm 0.424}$& $338.871$\\
    \bottomrule
    \end{tabular}}
    \label{table: riemann-res}
\end{table}

We also present some testing results in  \autoref{fig:riemann-ipr} for the IPR case 
and 
in \autoref{fig:riemann-hpr} for HPR case (LeBlanc-Sod). 
Results from the transformer show little or no overshooting or oscillation at the discontinuities. In comparison, the DeepONet exhibits slight overshoots and undershoots for the $u$ of IPR and $u$ and $p$ of HPR cases at the discontinuity locations.

\begin{figure}[tb!]
    \centering
    \begin{subfigure}{0.33\textwidth}
      \centering
      \includegraphics[width=\linewidth,height=\linewidth]{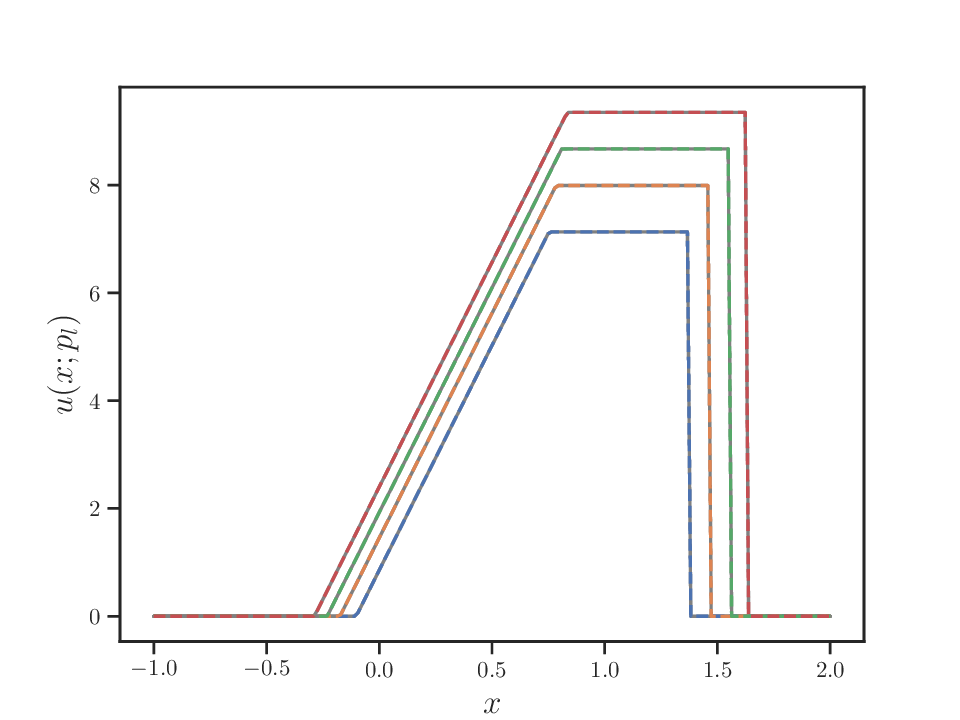}
      \caption{Transformer IPR $u$}
    \end{subfigure}%
    \begin{subfigure}{0.33\textwidth}
      \centering
      \includegraphics[width=\linewidth,height=\linewidth]{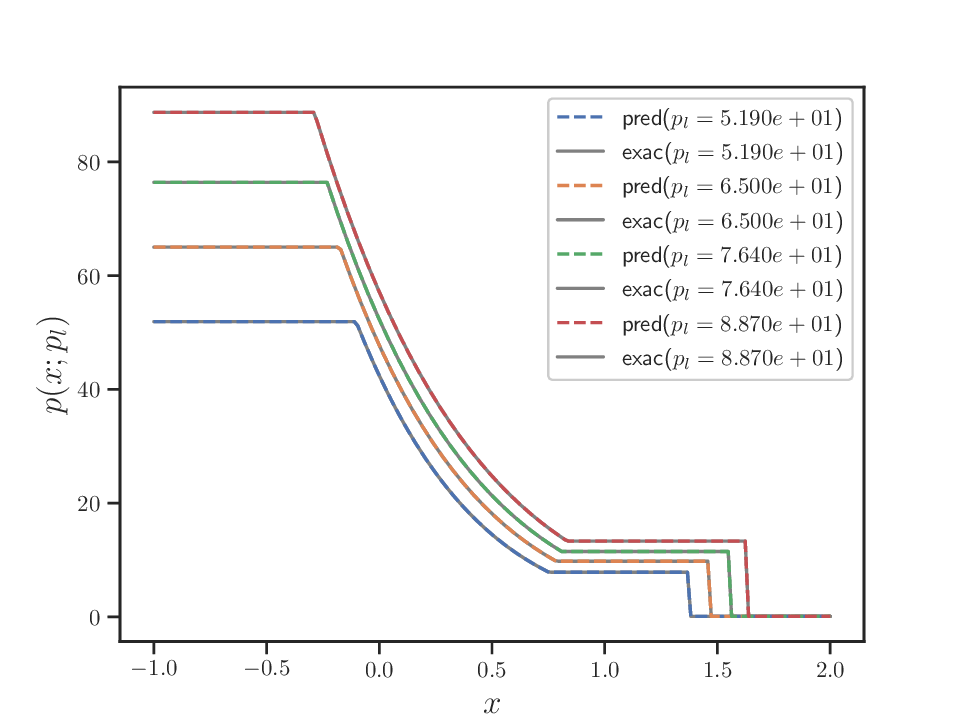}
      \caption{Transformer IPR $p$}
    \end{subfigure}
    \begin{subfigure}{0.33\textwidth}
      \centering
      \includegraphics[width=\textwidth,height=\linewidth]{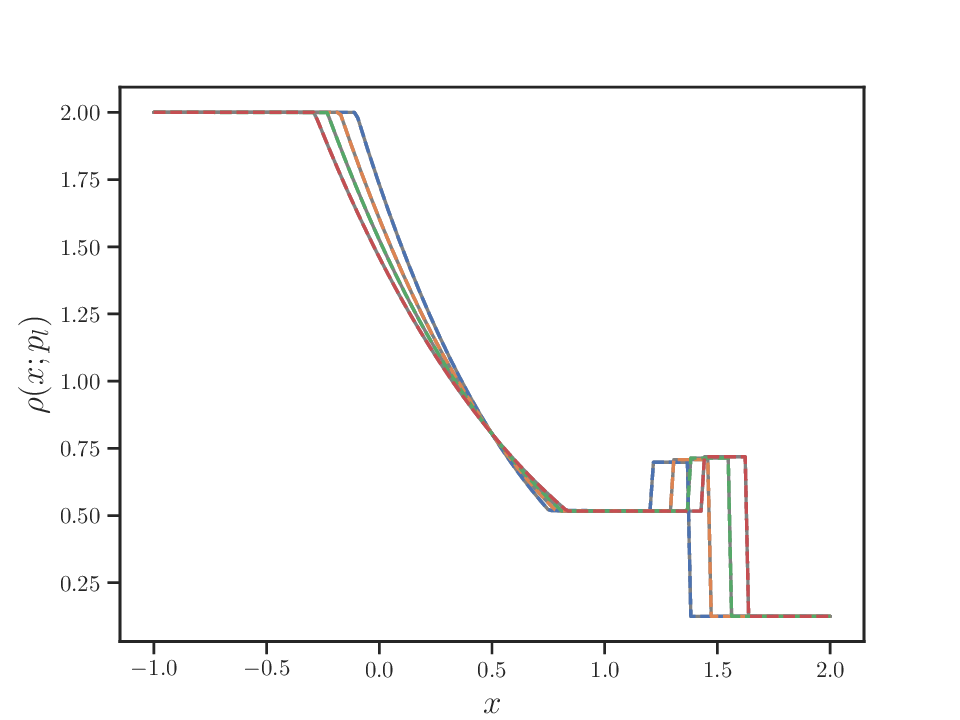}
      \caption{Transformer IPR $\rho$}
    \end{subfigure} 
    \\ 
    \begin{subfigure}{0.33\textwidth}
      \centering
      \includegraphics[width=\linewidth,height=\linewidth]{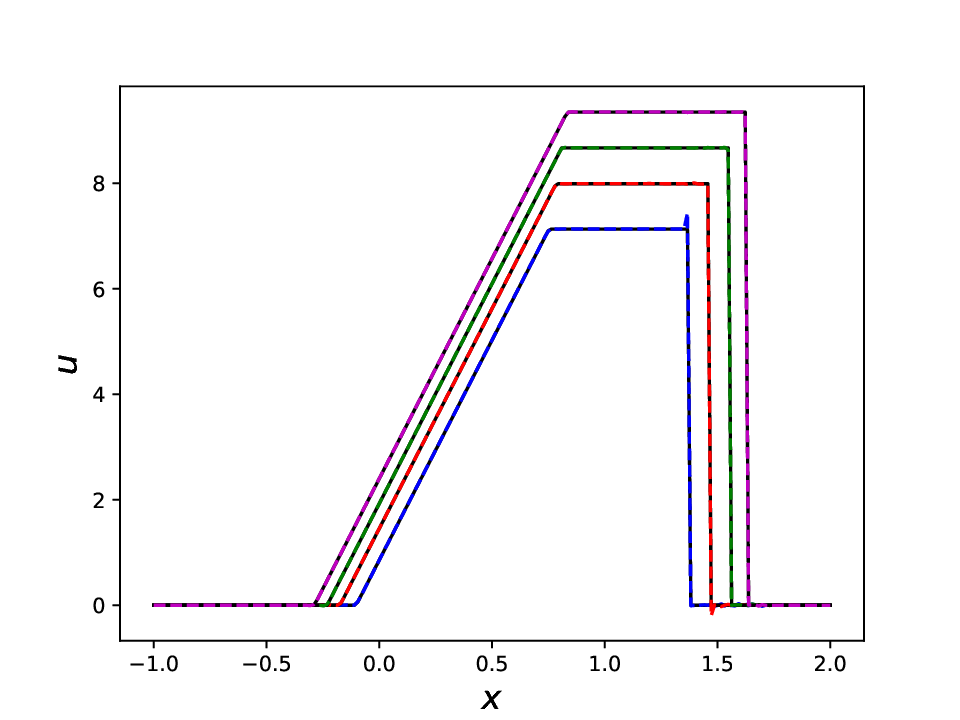}
      \caption{DeepONet IPR $u$}
    \end{subfigure}%
    \begin{subfigure}{0.33\textwidth}
      \centering
      \includegraphics[width=\linewidth,height=\linewidth]{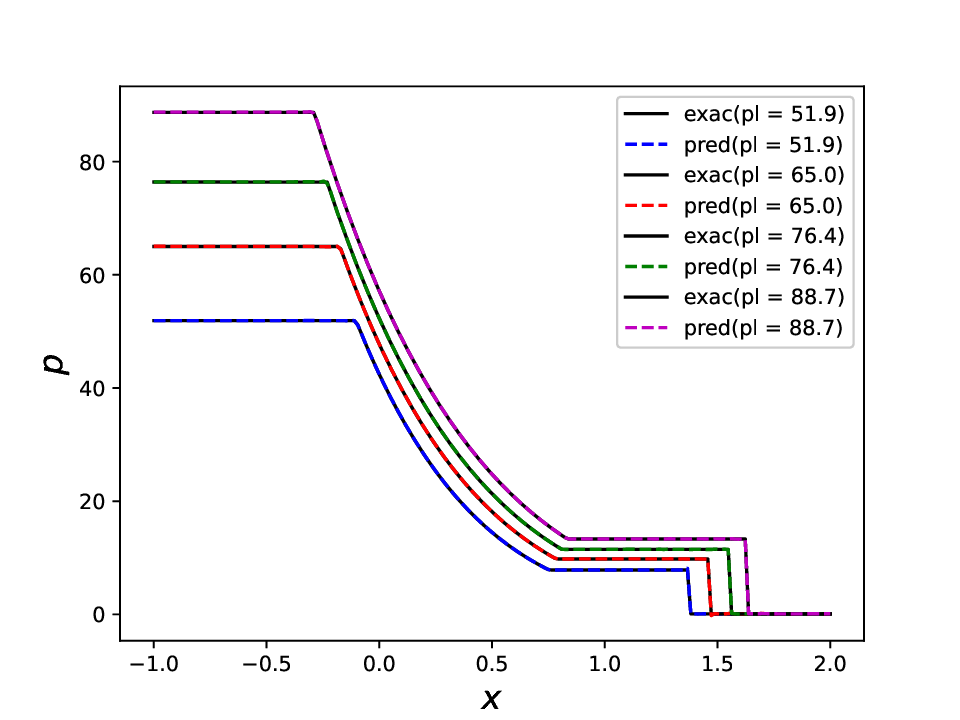}
      \caption{DeepONet IPR $p$}
    \end{subfigure}
    \begin{subfigure}{0.33\textwidth}
      \centering
      \includegraphics[width=\textwidth,height=\linewidth]{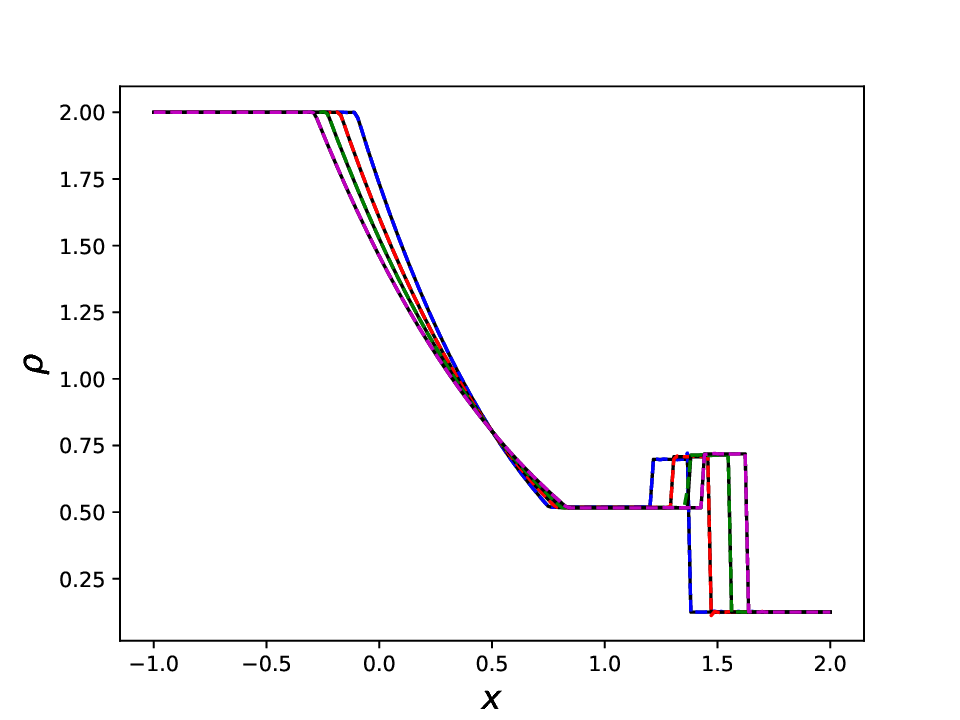}
      \caption{DeepONet IPR $\rho$}
    \end{subfigure} 
    \caption{Velocity, pressure, and density profiles for intermediate pressure ratio case of Sod's problem. The top row shows the inferences of the transformer for velocity, pressure, and density profiles for test datasets. The bottom row shows the results of DeepONet/RiemannOnet \cite{riemannonet}.}
    \label{fig:riemann-ipr}
\end{figure}

    

\begin{figure}[tb!]
    \centering
    \begin{subfigure}{0.33\textwidth}
      \centering
      \includegraphics[width=\linewidth,height=\linewidth]{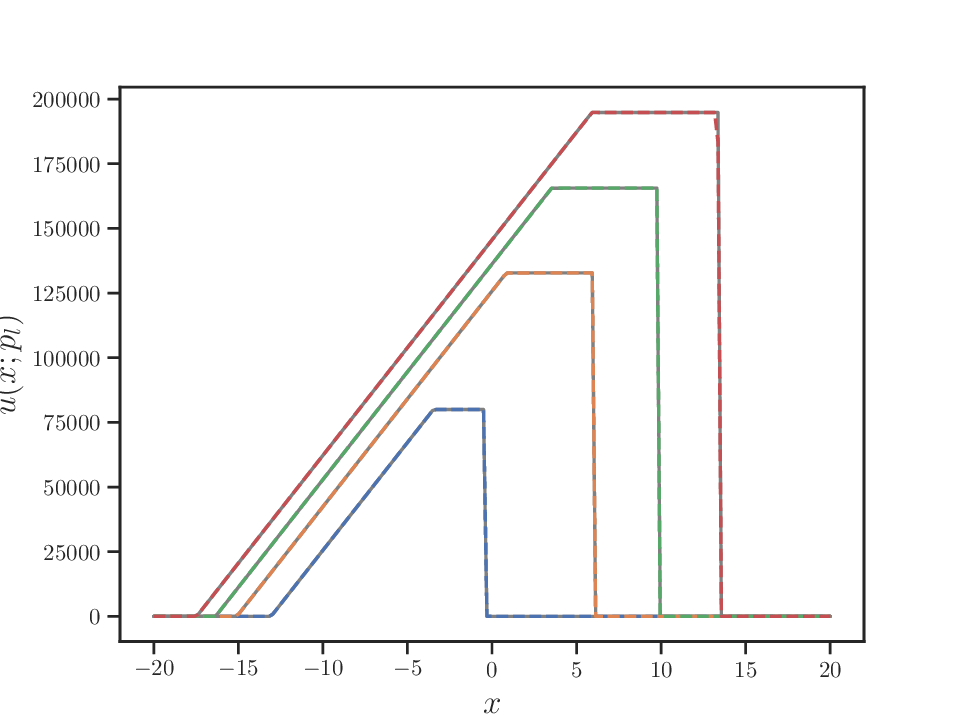}
      \caption{Transformer HPR $u$}
    \end{subfigure}%
    \begin{subfigure}{0.33\textwidth}
      \centering
      \includegraphics[width=\linewidth,height=\linewidth]{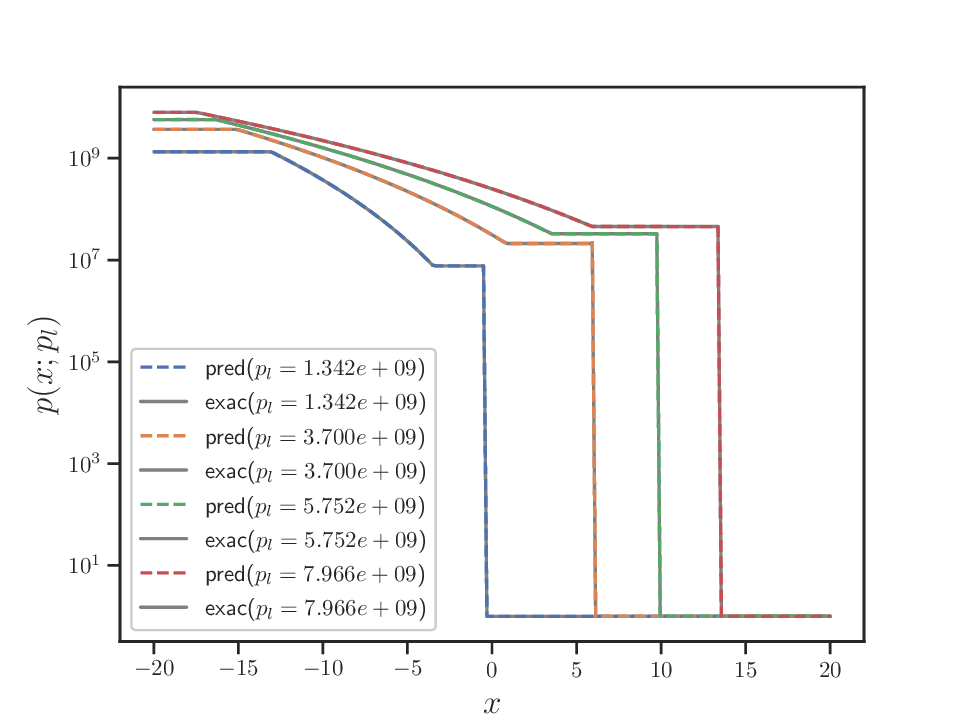}
      \caption{Transformer HPR $p$}
    \end{subfigure}
    \begin{subfigure}{0.33\textwidth}
      \centering
      \includegraphics[width=\textwidth,height=\linewidth]{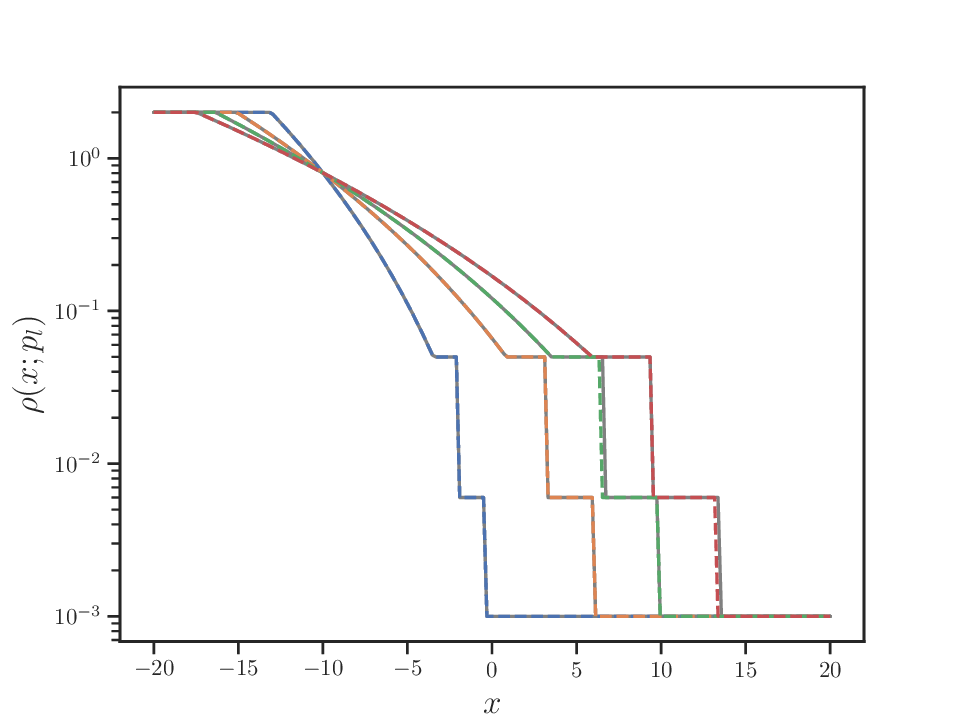}
      \caption{Transformer HPR $\rho$}
    \end{subfigure} 
    \\ 
    \begin{subfigure}{0.33\textwidth}
      \centering
      \includegraphics[width=\linewidth,height=\linewidth]{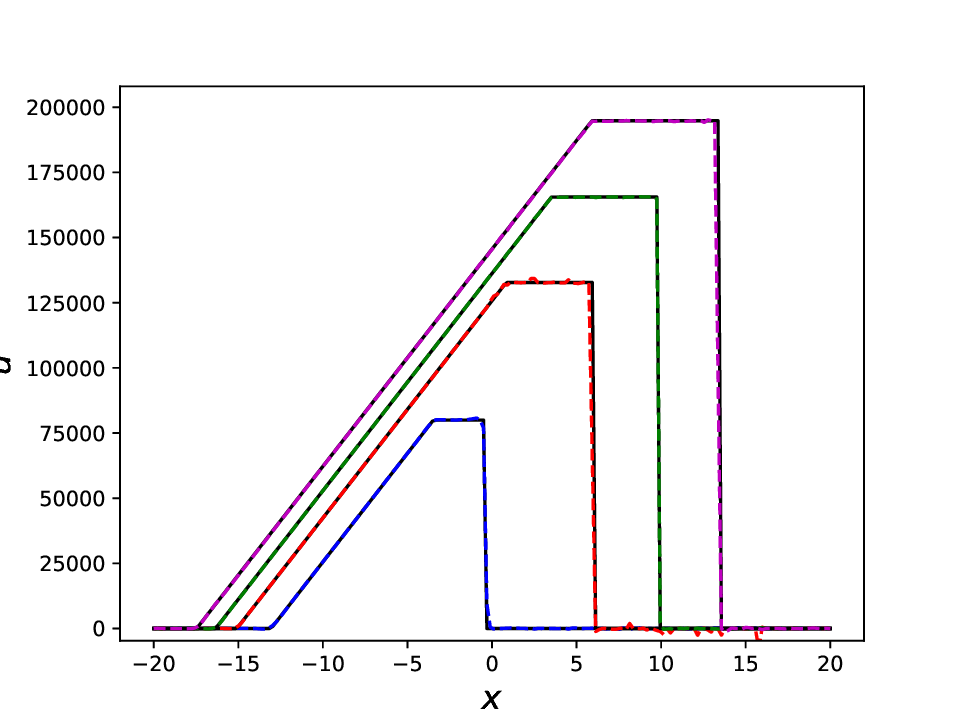}
      \caption{DeepONet HPR $u$}
    \end{subfigure}%
    \begin{subfigure}{0.33\textwidth}
      \centering
      \includegraphics[width=\linewidth,height=\linewidth]{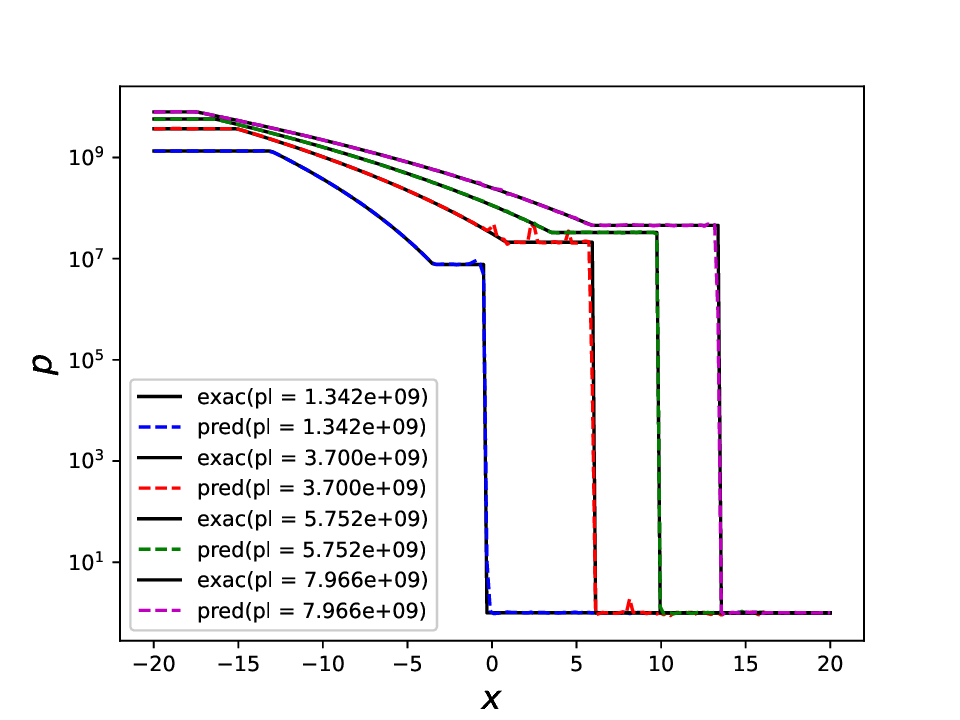}
      \caption{DeepONet HPR $p$}
    \end{subfigure}
    \begin{subfigure}{0.33\textwidth}
      \centering
      \includegraphics[width=\textwidth,height=\linewidth]{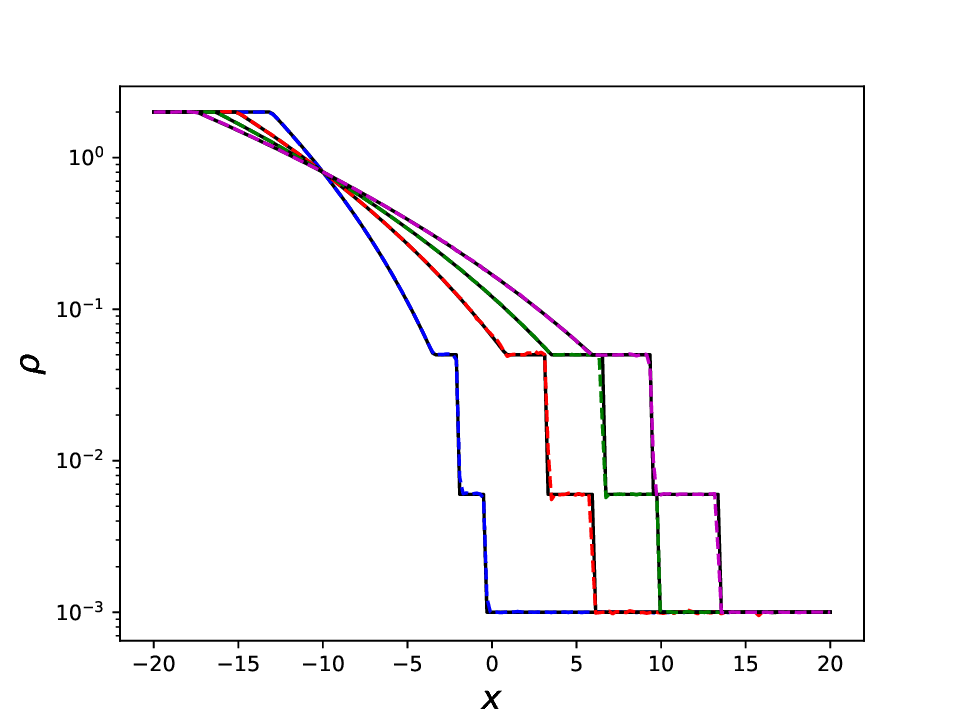}
      \caption{DeepONet HPR $\rho$}
    \end{subfigure} 
    \caption{Velocity, pressure, and density profiles for LeBlanc-Sod (high-pressure ratio case). The top row shows the inferences of the transformer for velocity, pressure, and density profiles for test datasets. The bottom row shows the results of DeepONet/RiemannOnet \cite{riemannonet}.}
    \label{fig:riemann-hpr}
\end{figure}


%
However, the transformer takes more time to train 
than the DeepONet of the form 
$\sum_{i=1}^p \mathcal{B}_i(p_l) \mathcal{T}_i(x) $, where $\mathcal{B}_i$ and $\mathcal{T}_i$ are both feedforward neural networks. 

\section{Summary}

In this work, we demonstrated the power of the attention mechanism in operator learning problems with low regularity.  Specifically, we investigated the approximation capabilities of these models on Sod's shock tube problem and two popular neuron models: the Izhikevich model and the tempered fractional LIF model. 
For Sod's problem, we compared the transformer and RiemannONet (a variant of DeepOnet) for the Riemann problem with various pressure ratios.
Our experiments revealed the remarkable learning capacity of the transformer, showing excellent accuracy across each system. However, the computational cost of transformers remains high, and alternative models such as state-space models (SSMs) should also be explored in the future. 


\section*{Acknowledgments}
This work was supported by the U.S. Army Research Laboratory
W911NF-22-2-0047 and by the MURI-AFOSR FA9550-20-1-0358. 
\appendix


\bibliographystyle{elsarticle-num} 
\bibliography{reference,ref,transformer}

\begin{thebibliography}{10}
\expandafter\ifx\csname url\endcsname\relax
  \def\url#1{\texttt{#1}}\fi
\expandafter\ifx\csname urlprefix\endcsname\relax\def\urlprefix{URL }\fi
\expandafter\ifx\csname href\endcsname\relax
  \def\href#1#2{#2} \def\path#1{#1}\fi

\bibitem{lu2021learning}
L.~Lu, P.~Jin, G.~Pang, Z.~Zhang, G.~E. Karniadakis, Learning nonlinear operators via deeponet based on the universal approximation theorem of operators, Nature machine intelligence 3~(3) (2021) 218--229.

\bibitem{lu2022comprehensive}
L.~Lu, X.~Meng, S.~Cai, Z.~Mao, S.~Goswami, Z.~Zhang, G.~E. Karniadakis, A comprehensive and fair comparison of two neural operators (with practical extensions) based on fair data, Computer Methods in Applied Mechanics and Engineering 393 (2022) 114778.

\bibitem{goswami2022Transferoperator}
S.~Goswami, K.~Kontolati, M.~D. Shields, G.~E. Karniadakis, Deep transfer operator learning for partial differential equations under conditional shift, Nature Machine Intelligence 4~(12) (2022) 1155--1164.

\bibitem{chen1995universal}
T.~Chen, H.~Chen, Universal approximation to nonlinear operators by neural networks with arbitrary activation functions and its application to dynamical systems, IEEE Transactions on Neural Networks 6~(4) (1995) 911--917.

\bibitem{li2020fourier}
Z.~Li, N.~Kovachki, K.~Azizzadenesheli, B.~Liu, K.~Bhattacharya, A.~Stuart, A.~Anandkumar, {F}ourier neural operator for parametric partial differential equations, arXiv:2010.08895 (2020).

\bibitem{kovachki2021universal}
N.~Kovachki, S.~Lanthaler, S.~Mishra, On universal approximation and error bounds for fourier neural operators, The Journal of Machine Learning Research 22~(1) (2021) 13237--13312.

\bibitem{fno-don-compare}
L.~Lu, X.~Meng, S.~Cai, Z.~Mao, S.~Goswami, Z.~Zhang, G.~E. Karniadakis, A comprehensive and fair comparison of two neural operators (with practical extensions) based on fair data, Computer Methods in Applied Mechanics and Engineering 393 (2022) 114778.

\bibitem{wang2021learning}
S.~Wang, H.~Wang, P.~Perdikaris, Learning the solution operator of parametric partial differential equations with physics-informed deeponets, Science advances 7~(40) (2021) eabi8605.

\bibitem{goswami2023physics}
S.~Goswami, A.~Bora, Y.~Yu, G.~E. Karniadakis, Physics-informed deep neural operator networks, in: Machine Learning in Modeling and Simulation: Methods and Applications, Springer, 2023, pp. 219--254.

\bibitem{luo2023efficient}
D.~Luo, T.~O'Leary-Roseberry, P.~Chen, O.~Ghattas, Efficient pde-constrained optimization under high-dimensional uncertainty using derivative-informed neural operators (2023).
\newblock \href {http://arxiv.org/abs/arXiv:2305.20053} {\path{arXiv:arXiv:2305.20053}}.

\bibitem{lu2021comprehensive}
L.~Lu, X.~Meng, S.~Cai, Z.~Mao, S.~Goswami, Z.~Zhang, G.~E. Karniadakis, A comprehensive and fair comparison of two neural operators (with practical extensions) based on {FAIR} data, Computer Methods in Applied Mechanics and Engineering 393 (2022) 114778.

\bibitem{Venturi2023svd-deeponet}
S.~Venturi, T.~Casey, \href{https://www.sciencedirect.com/science/article/pii/S0045782522006739}{Svd perspectives for augmenting deeponet flexibility and interpretability}, Computer Methods in Applied Mechanics and Engineering 403 (2023) 115718.
\newblock \href {https://doi.org/https://doi.org/10.1016/j.cma.2022.115718} {\path{doi:https://doi.org/10.1016/j.cma.2022.115718}}.
\newline\urlprefix\url{https://www.sciencedirect.com/science/article/pii/S0045782522006739}

\bibitem{lee2023training}
S.~Lee, Y.~Shin, On the training and generalization of deep operator networks, arXiv preprint arXiv:2309.01020 (2023).

\bibitem{zhang2023belnet}
Z.~Zhang, L.~Wing~Tat, H.~Schaeffer, Belnet: Basis enhanced learning, a mesh-free neural operator, Proceedings of the Royal Society A 479~(2276) (2023) 20230043.

\bibitem{franco2023mesh-informed}
N.~R. Franco, A.~Manzoni, P.~Zunino, Mesh-informed neural networks for operator learning in finite element spaces, Journal of Scientific Computing 97~(2) (2023) 35.

\bibitem{DengSLZK22}
B.~Deng, Y.~Shin, L.~Lu, Z.~Zhang, G.~E. Karniadakis, \href{https://www.sciencedirect.com/science/article/pii/S0893608022002349}{Approximation rates of {DeepONet}s for learning operators arising from advection–diffusion equations}, Neural Netw. 153 (2022) 411--426.
\newblock \href {https://doi.org/https://doi.org/10.1016/j.neunet.2022.06.019} {\path{doi:https://doi.org/10.1016/j.neunet.2022.06.019}}.
\newline\urlprefix\url{https://www.sciencedirect.com/science/article/pii/S0893608022002349}

\bibitem{standardattention}
A.~Vaswani, N.~Shazeer, N.~Parmar, J.~Uszkoreit, L.~Jones, A.~N. Gomez, L.~Kaiser, I.~Polosukhin, Attention is all you need, in: Proceedings of the 31st International Conference on Neural Information Processing Systems, NIPS'17, Curran Associates Inc., Red Hook, NY, USA, 2017, p. 6000–6010.

\bibitem{zappala2023neural}
E.~Zappala, A.~H. de~Oliveira~Fonseca, J.~O. Caro, D.~van Dijk, Neural integral equations, arXiv:2209.15190 (2023).

\bibitem{GENEVA2022272}
N.~Geneva, N.~Zabaras, \href{https://www.sciencedirect.com/science/article/pii/S0893608021004500}{Transformers for modeling physical systems}, Neural Networks 146 (2022) 272--289.
\newblock \href {https://doi.org/https://doi.org/10.1016/j.neunet.2021.11.022} {\path{doi:https://doi.org/10.1016/j.neunet.2021.11.022}}.
\newline\urlprefix\url{https://www.sciencedirect.com/science/article/pii/S0893608021004500}

\bibitem{li2023-transformerPDE}
Z.~Li, K.~Meidani, A.~B. Farimani, \href{https://openreview.net/forum?id=EPPqt3uERT}{Transformer for partial differential equations{\textquoteright} operator learning}, Transactions on Machine Learning Research (2023).
\newline\urlprefix\url{https://openreview.net/forum?id=EPPqt3uERT}

\bibitem{li2024latent}
Z.~Li, S.~Patil, F.~Ogoke, D.~Shu, W.~Zhen, M.~Schneier, J.~R. Buchanan~Jr, A.~B. Farimani, Latent neural pde solver: a reduced-order modelling framework for partial differential equations, arXiv preprint arXiv:2402.17853 (2024).

\bibitem{liu2024mitigating}
X.~Liu, B.~Xu, S.~Cao, L.~Zhang, Mitigating spectral bias for the multiscale operator learning, Journal of Computational Physics 506 (2024) 112944.

\bibitem{ovadia2023vito}
O.~Ovadia, A.~Kahana, P.~Stinis, E.~Turkel, G.~E. Karniadakis, Vito: Vision transformer-operator, arXiv preprint arXiv:2303.08891 (2023).

\bibitem{ovadia2023ditto}
O.~Ovadia, E.~Turkel, A.~Kahana, G.~E. Karniadakis, Ditto: Diffusion-inspired temporal transformer operator, arXiv preprint arXiv:2307.09072 (2023).

\bibitem{guo2022transformer}
R.~Guo, S.~Cao, L.~Chen, Transformer meets boundary value inverse problems, in: The Eleventh International Conference on Learning Representations, 2022.

\bibitem{riemannonet}
A.~Peyvan, V.~Oommen, A.~D. Jagtap, G.~E. Karniadakis, Riemannonets: Interpretable neural operators for riemann problems, Computer Methods in Applied Mechanics and Engineering 426 (2024) 116996.

\bibitem{cordonnier2020.RelationshipSelfAttentionConvolutional}
J.-B. Cordonnier, A.~Loukas, M.~Jaggi, On the {{Relationship}} between {{Self-Attention}} and {{Convolutional Layers}}, in: International {{Conference}} on {{Learning Representations}}, 2020.

\bibitem{pmlr-v202-takakura23a}
S.~Takakura, T.~Suzuki, \href{https://proceedings.mlr.press/v202/takakura23a.html}{Approximation and estimation ability of transformers for sequence-to-sequence functions with infinite dimensional input}, in: A.~Krause, E.~Brunskill, K.~Cho, B.~Engelhardt, S.~Sabato, J.~Scarlett (Eds.), Proceedings of the 40th International Conference on Machine Learning, Vol. 202 of Proceedings of Machine Learning Research, PMLR, 2023, pp. 33416--33447.
\newline\urlprefix\url{https://proceedings.mlr.press/v202/takakura23a.html}

\bibitem{Yun2020Are}
C.~Yun, S.~Bhojanapalli, A.~S. Rawat, S.~Reddi, S.~Kumar, \href{https://openreview.net/forum?id=ByxRM0Ntvr}{Are transformers universal approximators of sequence-to-sequence functions?}, in: International Conference on Learning Representations, 2020.
\newline\urlprefix\url{https://openreview.net/forum?id=ByxRM0Ntvr}

\bibitem{chen1995approximation}
T.~Chen, H.~Chen, Approximation capability to functions of several variables, nonlinear functionals, and operators by radial basis function neural networks, IEEE Transactions on Neural Networks 6~(4) (1995) 904--910.

\bibitem{lanthaler2021error}
S.~Lanthaler, S.~Mishra, G.~E. Karniadakis, Error estimates for {DeepOnet}s: A deep learning framework in infinite dimensions (a2021).
\newblock \href {http://arxiv.org/abs/arXiv:2102.09618} {\path{arXiv:arXiv:2102.09618}}.

\bibitem{chen1993approximations}
T.~Chen, H.~Chen, Approximations of continuous functionals by neural networks with application to dynamic systems, IEEE Transactions on Neural Networks 4~(6) (1993) 910--918.

\bibitem{holden2015front}
H.~Holden, N.~H. Risebro, Front tracking for hyperbolic conservation laws, second edition Edition, Springer, 2015.

\bibitem{Mhaskar23-localapproximation}
H.~Mhaskar, \href{https://www.sciencedirect.com/science/article/pii/S1063520323000052}{Local approximation of operators}, Applied and Computational Harmonic Analysis 64 (2023) 194--228.
\newblock \href {https://doi.org/https://doi.org/10.1016/j.acha.2023.01.004} {\path{doi:https://doi.org/10.1016/j.acha.2023.01.004}}.
\newline\urlprefix\url{https://www.sciencedirect.com/science/article/pii/S1063520323000052}

\bibitem{izhikevich2003simple}
E.~Izhikevich, Simple model of spiking neurons, IEEE Transactions on Neural Networks 14~(6) (2003) 1569--1572.
\newblock \href {https://doi.org/10.1109/TNN.2003.820440} {\path{doi:10.1109/TNN.2003.820440}}.

\bibitem{lif}
A.~Mabrouk, M.~E. Fouda, A.~Eltawil, On numerical approximations of fractional-order spiking neuron models, Communications in Nonlinear Science and Numerical Simulation 105 (2022) 106078.
\newblock \href {https://doi.org/10.1016/j.cnsns.2021.106078} {\path{doi:10.1016/j.cnsns.2021.106078}}.

\bibitem{fanhai}
Z.~Yang, F.~Zeng, A corrected l1 method for a time-fractional subdiffusion equation, Journal of Scientific Computing 95~(3) (2023) 85.

\bibitem{Cao2021transformer}
S.~Cao, Choose a transformer: Fourier or galerkin, in: M.~Ranzato, A.~Beygelzimer, Y.~Dauphin, P.~Liang, J.~W. Vaughan (Eds.), Advances in Neural Information Processing Systems, Vol.~34, Curran Associates, Inc., 2021, pp. 24924--24940.

\bibitem{OFormer}
Z.~Li, K.~Meidani, A.~B. Farimani, \href{https://openreview.net/forum?id=EPPqt3uERT}{Transformer for partial differential equations{\textquoteright} operator learning}, Transactions on Machine Learning Research (2023).
\newline\urlprefix\url{https://openreview.net/forum?id=EPPqt3uERT}

\bibitem{chen2021crossvit}
C.-F.~R. Chen, Q.~Fan, R.~Panda, Crossvit: Cross-attention multi-scale vision transformer for image classification, in: 2021 IEEE/CVF International Conference on Computer Vision (ICCV), 2021, pp. 347--356.
\newblock \href {https://doi.org/10.1109/ICCV48922.2021.00041} {\path{doi:10.1109/ICCV48922.2021.00041}}.

\bibitem{gegelu}
N.~Shazeer, \href{https://arxiv.org/abs/2002.05202}{{GLU} variants improve transformer}, CoRR abs/2002.05202 (2020).
\newblock \href {http://arxiv.org/abs/2002.05202} {\path{arXiv:2002.05202}}.
\newline\urlprefix\url{https://arxiv.org/abs/2002.05202}

\bibitem{adam}
D.~P. Kingma, J.~Ba, \href{http://arxiv.org/abs/1412.6980}{Adam: {A} method for stochastic optimization}, in: Y.~Bengio, Y.~LeCun (Eds.), 3rd International Conference on Learning Representations, {ICLR} 2015, San Diego, CA, USA, May 7-9, 2015, Conference Track Proceedings, 2015.
\newline\urlprefix\url{http://arxiv.org/abs/1412.6980}

\bibitem{onecyclelr}
L.~N. Smith, N.~Topin, \href{https://openreview.net/forum?id=H1A5ztj3b}{Super-convergence: Very fast training of residual networks using large learning rates} (2018).
\newline\urlprefix\url{https://openreview.net/forum?id=H1A5ztj3b}

\bibitem{lion-optimizer}
X.~Chen, C.~Liang, D.~Huang, E.~Real, K.~Wang, Y.~Liu, H.~Pham, X.~Dong, T.~Luong, C.-J. Hsieh, Y.~Lu, Q.~V. Le, \href{https://arxiv.org/abs/2302.06675}{Symbolic discovery of optimization algorithms} (2023).
\newline\urlprefix\url{https://arxiv.org/abs/2302.06675}

\end{thebibliography}
\end{document}